\documentclass{article}

\PassOptionsToPackage{numbers, compress}{natbib}
\usepackage[preprint]{neurips_2024}

\usepackage[utf8]{inputenc} % allow utf-8 input
\usepackage[T1]{fontenc}    % use 8-bit T1 fonts
\usepackage{hyperref}       % hyperlinks
\usepackage{url}            % simple URL typesetting
\usepackage{booktabs}       % professional-quality tables
\usepackage{amsfonts}       % blackboard math symbols
\usepackage{nicefrac}       % compact symbols for 1/2, etc.
\usepackage{microtype}      % microtypography
\usepackage{xcolor}         % colors

\usepackage{epsfig}
\usepackage{graphicx}
\usepackage{amsmath}
\usepackage{amssymb}
\usepackage{booktabs}
\usepackage{multirow}
\usepackage{tabularx}
\usepackage{enumitem}
\usepackage{gensymb}
\usepackage{colortbl}
\usepackage{caption}
\usepackage{makecell}
\usepackage{wrapfig}
\usepackage{amssymb}% http://ctan.org/pkg/amssymb
\usepackage{pifont}% http://ctan.org/pkg/pifont
\definecolor{mycyan}{cmyk}{.1,0,0,0}
\newcommand{\cmark}{\ding{51}}%
\newcommand{\cmarkg}{\textcolor{lightgray}{\ding{51}}}%
\newcommand{\xmarkg}{\textcolor{lightgray}{\ding{55}}}%
\newcommand{\mypara}[1]{\vspace{1mm}\noindent\textbf{#1}}
\usepackage[capitalize]{cleveref}
\newcommand{\name}{BEV-TSR}
\newcommand{\nusr}{nuScenes-Retrieval}

\usepackage{colortbl}
\definecolor{mygray}{gray}{.96}
\definecolor{mypink}{rgb}{.99,.91,.95}
\definecolor{mycyan}{cmyk}{.1,0,0,0}

\title{\name{}: Text-Scene Retrieval in BEV Space for Autonomous Driving}

\author{Tao Tang$^{2,1}$$^*$, Dafeng Wei$^{1}$$^*$, Zhengyu Jia$^{1}$$^*$, Tian Gao$^{1}$$^*$, \\ Changwei Cai$^{1}$$^{**}$, Chengkai Hou$^{1}$$^{**}$, Peng Jia$^{1}$, Kun Zhan$^{1}$, Haiyang Sun$^{1}$, \\ Jingchen Fan$^{1}$, Yixing Zhao$^{1}$, Fu Liu$^{1}$, Xiaodan Liang$^{2}$, Xianpeng Lang$^{1}$, Yang Wang$^{1}$$^{\P}$\\
\fontsize{11pt}{\baselineskip}\selectfont
$^1$ Li Auto Inc. \\
\fontsize{11pt}{\baselineskip}\selectfont
$^2$ Shenzhen Campus of Sun Yat-sen University \\
\tt\small \{weidafeng, jiazhengyu, gaotian, wangyang25\}@lixiang.com}

\begin{document}

\renewcommand{\thefootnote}{}
\footnotetext{ $^*$ Equal contribution}
\footnotetext{ $^{**}$ Equal contribution}
\footnotetext{ $^{\P}$ Corresponding Author}

% \maketitle
% \begin{figure}[t] %H为当前位置，!htb为忽略美学标准，htbp为浮动图形
% \centering %图片居中
% % \vspace{-3em}
% \includegraphics[width=0.8\linewidth]{Figure/teaser.pdf} %插入图片，[]中设置图片大小，{}中是图片文件名
% \caption{\textbf{\name{}, the first BEV retrieval method retrieves corner cases on autonomous driving.} In contrast to 2D image retrieval, \name{} allows semantic retrieval related to complex global features in the context of BEV features, which enables spa conquers. Meanwhile, \name{} uses a Large Language Model (LLM) to enhance the model's ability to understand complex descriptions in the retrieved text. \Tao{traiditional ITR vs our AD scene Retrieval;add performance}} %最终文档中希望显示的图片标题

% \vspace{-1.0em}
% \label{fig: teaser} %用于文内引用的标签
% \end{figure}

% \twocolumn[{ 
% \maketitle
%   \centering
%   % \vspace{-0.2cm}
%   \includegraphics[width=1.0\textwidth]{Figure/teaser.pdf}
%   \vspace{-6mm}
%   \captionof{figure}{\textbf{\name{}, the first BEV retrieval method retrieves corner cases on autonomous driving.} In contrast to 2D image retrieval, \name{} allows semantic retrieval related to complex global features in the context of BEV features, which enables spa conquers. Meanwhile, \name{} uses a Large Language Model (LLM) to enhance the model's ability to understand complex descriptions in the retrieved text. \Tao{traiditional ITR vs our AD scene Retrieval;add performance}
%    }
% \label{fig:fig1}
% \vspace{3mm}
% }]

\maketitle
\begin{center}
    \centering
    \captionsetup{type=figure}
    \begin{center}
        \includegraphics[width=1\textwidth]{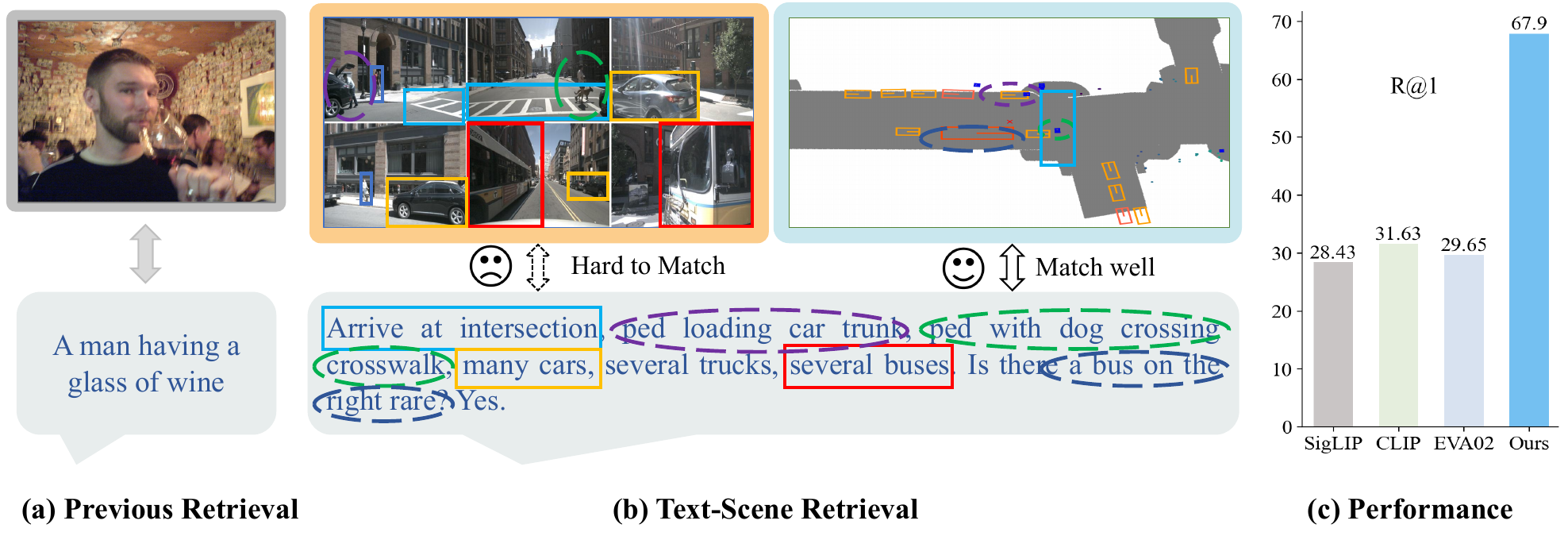}
    \end{center}
    \captionof{figure}{
        \textbf{(a)} Existing methods are primarily tailored for simple retrieval scenarios. \textbf{(b)} On the contrary, autonomous driving scenarios are challenging with numerous traffic participants and road elements. Then the BEV space offers a clearer global context of the scene than the previous image space, which aligns well with the textual query and serves as an ideal retrieval space.
        \textbf{(c)} To this end, we propose the novel \name{} framework for text-scene retrieval in autonomous driving, which retrieves scenes in BEV space and demonstrates a significant capability to retrieve traffic scenarios. }
\label{fig:teaser}
\end{center}
\begin{abstract}
The rapid development of the autonomous driving industry has led to a significant accumulation of autonomous driving data. Consequently, there comes a growing demand for retrieving data to provide specialized optimization.
However, directly applying previous image retrieval methods faces several challenges, such as the lack of global feature representation and inadequate text retrieval ability for complex driving scenes.
To address these issues, firstly, we propose the \textbf{\name{}} framework which leverages descriptive text as an input to retrieve corresponding scenes in the Bird's Eye View (BEV) space. 
Then to facilitate complex scene retrieval with extensive text descriptions, we employ a large language model (LLM) to extract the semantic features of the text inputs and incorporate knowledge graph embeddings to enhance the semantic richness of the language embedding.
To achieve feature alignment between the BEV feature and language embedding, we propose Shared Cross-modal Embedding with a set of shared learnable embeddings to bridge the gap between these two modalities, and employ a caption generation task to further enhance the alignment.
Furthermore, there lack of well-formed retrieval datasets for effective evaluation. 
To this end, we establish a multi-level retrieval dataset, \nusr{}, based on the widely adopted nuScenes dataset.
% Furthermore, the autonomous driving community lacks well-formed retrieval datasets for effective evaluation. 
% To this end, we develop a toolkit for constructing multi-level retrieval datasets using detection datasets, and we showcase the effectiveness of our toolkit by establishing \nusr{} on the widely adopted nuScenes dataset.
Experimental results on the multi-level \nusr{} show that \name{} achieves state-of-the-art performance, e.g., 85.78\% and 87.66\% top-1 accuracy on scene-to-text and text-to-scene retrieval respectively.
Codes and datasets will be available.

\end{abstract}    
\section{Introduction}
\label{sec:intro}
% 开头有点长，有点拗口
The past few years have witnessed rapid advancements in the autonomous driving industry~\cite{ma2022vision, li2023delving}, which transitioned from a phase of data scarcity to one where data is abundant, owing to the substantial data generated by both data collection vehicles and crowdsourcing vehicles. 
% The industry has transitioned from a phase of data scarcity to one of abundant data, as both data collection vehicles and crowdsourcing vehicles have generated substantial amounts of data. 
However, the mere accumulation of uniformly distributed data is no longer sufficient to achieve significant improvements~\cite{long2022retrieval}.
For example, if we aim to ensure effective performance on rural roads, it is imperative to retrieve a sufficient amount of rural road data to fine-tune the models. 
Therefore, data mining has become an essential working schema to provide specialized optimization for autonomous driving models, and a well-designed retrieval method plays a crucial role in the closed-loop data-driven pipeline of autonomous driving data~\cite{fu2024limsim++, li2024data}.

% The past few years have witnessed rapid advancements in the autonomous driving industry~\cite{ma2022vision, li2023delving}, 
% with onboard autonomous vehicles equipped with multiple sensor inputs becoming increasingly prevalent in traffic scenarios. 
% % with the increasing prevalence of onboard autonomous vehicles equipped with multiple sensor inputs in various traffic scenarios.
% This transition has led the industry from a phase of data scarcity to one where data is abundant, owing to the substantial amount of data generated by both data collection vehicles and crowdsourcing vehicles. 
% % The industry has transitioned from a phase of data scarcity to one of abundant data, as both data collection vehicles and crowdsourcing vehicles have generated substantial amounts of data. 
% However, the mere accumulation of uniformly distributed data is no longer sufficient to achieve significant improvements~\cite{long2022retrieval}.
% For example, if we aim to ensure effective performance on rural roads, it is imperative to retrieve a sufficient amount of rural road data to fine-tune the models. 
% Therefore, data mining has become an essential working schema to provide specialized optimization for autonomous driving models, and a well-designed retrieval method plays a crucial role in the closed-loop data-driven pipeline of autonomous driving data~\cite{fu2024limsim++, li2024data}.

On the other hand, cross-modal image-text retrieval (ITR) has been a longstanding research task and presents a significant advancement over the past few years as to the prosperity of deep models for language and vision~\cite{cao2022image, ma2021image}.
However, as shown in \cref{fig:teaser}, despite significant recent progress in the field of ITR, when applied to challenging autonomous driving scenarios, modern retrieval methods often struggle to achieve satisfactory results. 
% The complexity of autonomous driving scenes poses unique challenges that previous methods were not designed to address.
While previous methods~\cite{fang2023eva, li2023blip, zhai2023sigmoid} have been successful in handling simpler tasks such as identifying objects like "a brown dog" or more complex descriptions like "a man having a glass of wine," autonomous driving datasets contain numerous traffic participants and road elements, e.g., 
%"wait at an intersection, nature, scooters, pedestrians on the sidewalk, pedestrians with umbrellas, parked bicycles, taxis, jaywalkers, delivery scooters, turning right, multiple cars, one truck, one bicycle, multiple motorcycles, multiple pedestrians."
"arrive at intersection, ped loading car trunk, ped with dog crossing crosswalk, many cars, several buses."
These complex traffic scenes present significant challenges for retrieval models in autonomous driving: models need to possess a comprehensive understanding of the global context of traffic scenes and the ability to comprehend complex and lengthy textual inputs. 
These requirements go beyond the capabilities of existing retrieval methods.

In this paper, we propose \name{} framework to tackle the barriers from the main components of the retrieval pipeline.
The traditional ITR paradigm involves two main steps: (1) extracting the representation of images and sentences; and (2) aligning the cross-model representations with similar semantics.
For feature extraction, for the scene image representation, we propose to retrieve corresponding scenes in the Bird's Eye View (BEV) space. As shown in \cref{fig:teaser} (b), the image space often presents scattered and truncated elements of traffic scenes, e.g., the two images highlighted within the red box only display partial views of the bus. While the BEV space offers a clearer global context of the scene, which aligns well with the textual query and serves as an ideal retrieval space.
For text sentence representation, to facilitate complex scene retrieval with extensive text descriptions, we utilize a large language model (LLM) to extract semantic features from textual inputs. 
% Moreover, we enhance the semantic richness and diversity of the language embedding by incorporating semi-structured information from a knowledge graph.
Moreover, we incorporate learned knowledge graph embeddings of autonomous driving which further enhances the understanding and comprehension of the textual elements. 
% By combining these two components, our \name{} framework achieves a higher level of semantic richness and diversity in the language embedding, enabling more effective and accurate retrieval of scenes in autonomous driving scenarios.
Then, for feature alignment, with the BEV feature and language embedding in different feature spaces, we propose the Shared Cross-modal Embedding, which utilizes a set of shared learnable embeddings to bridge the gap between the two modalities. Additionally, we employ a caption generation task to further enhance the alignment.
% Furthermore, the autonomous driving community lacks well-structured retrieval datasets for effective evaluation. To address this, we have developed a toolkit for constructing multi-level retrieval datasets using detection datasets. To demonstrate the effectiveness of our toolkit, we establish the \nusr{} dataset based on the widely adopted nuScenes dataset.
Furthermore, there lacks well-structured retrieval datasets for effective evaluation. To address this, we establish a multi-level retrieval dataset, \nusr{}, based on the widely adopted nuScenes dataset.
Experimental results on the multi-level \nusr{} demonstrate that \name{} achieves significant advancements, e.g., 85.78\% and 87.66\% top-1 accuracy on scene-to-text and text-to-scene retrieval respectively.

Our main contributions can be summarized as follows:
\begin{enumerate}
  \item We propose the novel \name{} framework for text-scene retrieval in autonomous driving, which retrieves scenes in BEV space and demonstrates a significant capability to understand the global context and retrieve complex traffic scenarios.
  \item We leverage an LLM and incorporate knowledge graph embeddings to comprehensively understand complex textual descriptions, offering a higher level of semantic richness in language embedding.
  \item We propose Shared Cross-modal Embedding with a set of shared learnable embeddings to align the cross-model features, and employ a caption generation task to further enhance the alignment.
  \item We establish a multi-level retrieval dataset, \nusr{}, based on the nuScenes dataset, on which our \name{} achieves state-of-the-art performance, e.g., 85.78\% and 87.66\% top-1 accuracy on scene-to-text and text-to-scene retrieval respectively.
\end{enumerate}

\section{Related work}

\mypara{Image-Text Retrieval.} 
Image-Text Retrieval (ITR) is a fundamental cross-modal task in computer vision, whose main challenge lies in learning a shared representation of images and texts and accurately measuring their similarity~\cite{cao2022image}. 
Traditional methods for text-to-image retrieval have typically relied on convolutional neural networks (CNNs) as independent encoders to produce representations to measure the similarities between images and textual content~\cite{dong2014learning, noh2017large, radenovic2018fine}.
Recent years have witnessed a surge in the adoption of transform-based models and large-scale language-image pre-training~\cite{radford2021learning, fang2023eva, li2023blip, zhai2023sigmoid},
% leading to significant advancements, 
which achieve state-of-the-art performance across various text-to-image benchmark tasks.
However, existing methods are primarily tailored for simple retrieval scenarios with simple text inputs, focusing on single-image retrieval, and are evaluated on datasets such as MSCOCO~\cite{lin2014microsoft} and Flickr30k~\cite{plummer2015flickr30k}.
% , which provide only one relevant ground truth image per textual query. 
The potential applications to large-scale complex scenarios remain largely unexplored.
In this work, we first delve into the text-scene retrieval of autonomous driving. We find that modern retrieval methods often struggle to achieve satisfactory results when applied to challenging autonomous driving scenarios, as they cannot comprehensively understand the global context of traffic scenes and comprehend complex and lengthy textual inputs. 
To this end, we propose the \name{} framework which incorporates several elaborate modules for feature extraction and alignment, to achieve accurate text-scene retrieval of autonomous driving.

\mypara{BEV Space Learning.}
In recent years, learning powerful representations in Bird’s Eye View (BEV) space has been a growing trend and garnered significant attention from both industry and academia~\cite{ma2022vision, li2023delving}. 
BEV approaches have gained popularity in various aspects of autonomous driving, including detection~\cite{ wang2023StreamPETR, park2022time, park2022time}, segmentation~\cite{liang2022bevfusion,liu2023bevfusion, chen2022persformer}, tracking~\cite{lin2023sparse4d}, and occupancy forecasting~\cite{zhang2023occformer, wang2023openoccupancy, tong2023scene}. 
% The core problems for BEV space learning lie in how to reconstruct the BEV features from the perspective view.
The main challenge in learning within the BEV space is the reconstruction of BEV features from the perspective view. 
Following LSS~\cite{philion2020lift}, some methods like BEVDet series~\cite{huang2021bevdet, li2023bevdepth, li2023bevstereo} predict a distribution over depth bins and lift weighted image features onto BEV.
Another branch of work follows DETR3D~\cite{wang2022detr3d} and adopts BEV queries and projects them to get image features. BEVFormer~\cite{li2022bevformer} introduces sequential temporal modeling into multi-view 3D object detection and applies temporal self-attention.
PETR~\cite{liu2022petr} proposes 3D position embedding in the global system and then conducts global cross-attention to update the queries.
These methods have gained popularity due to the advantages of BEV space learning, such as providing an intuitive representation of the world and enabling the representation of objects in BEV, which is most desirable for subsequent modules in planning and control. These characteristics naturally make BEV space an optimal space for text-scene retrieval. Therefore, in this paper, we propose \name{}, the first text-scene retrieval method in the BEV space.

\section{\name{}}
% In this section, we discuss main structure of \name{}, a methodology of text-to-BEV contrastive learning retrieval. We obtain the method of utilizing the pre-trained BEV encoder weight for retrieval tasks and apply cross-modal interacting strategy with the language representation, which fused by text description and knowledge graph embedding.

In this section, we introduce our proposed \name{} in detail.  We first give a brief problem definition and an overview of the framework in \cref{subsec:problem}. Then, we subsequently delve into our key contributions on the feature extraction and feature alignment in \cref{subsec:extract} and \cref{subsec:alignment}.
% subsequently delve into 

\begin{figure*}[t] %H为当前位置，!htb为忽略美学标准，htbp为浮动图形
% \vspace{-1.2cm}
\centering %图片居中
\includegraphics[width=1\textwidth]{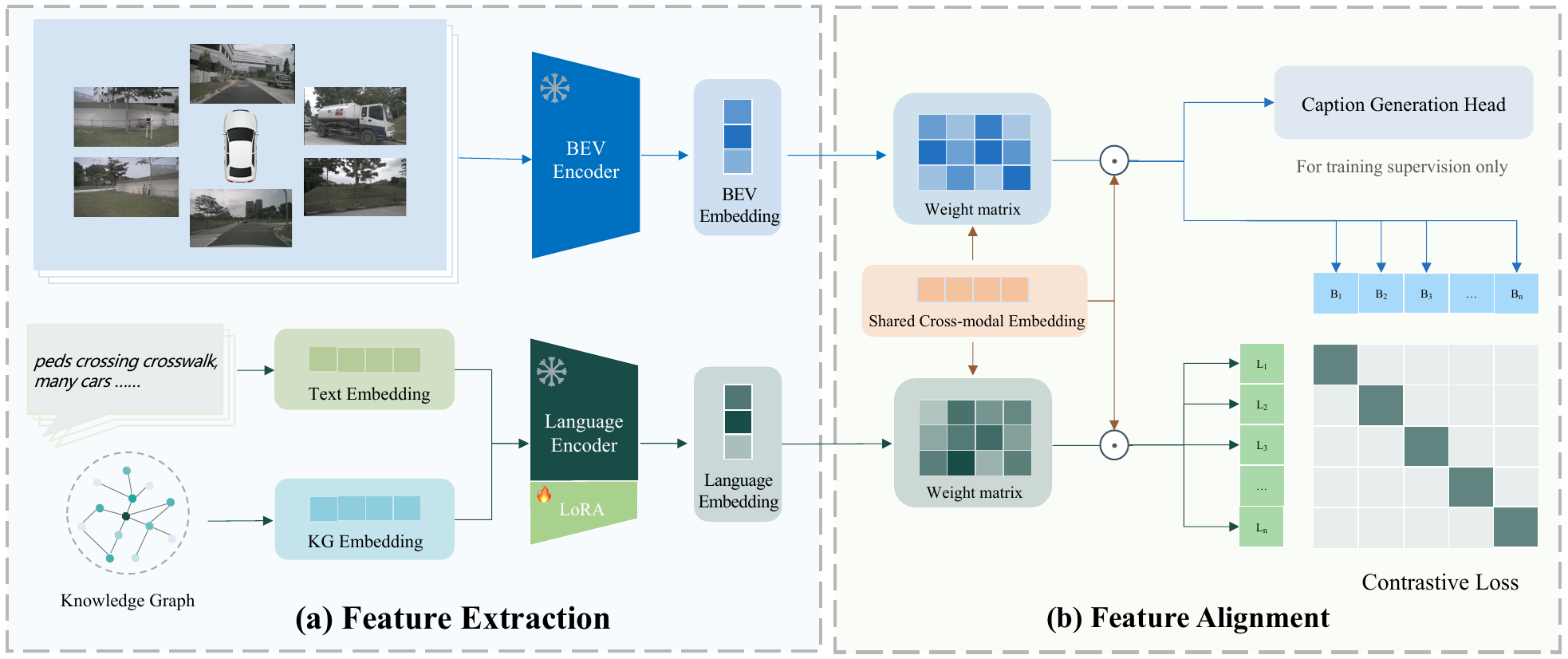} 
% \vspace{-0.6cm}
\caption{\textbf{Overall framework of \name{}.}
\textbf{(a) Feature Extraction.} For the visual branch, the BEV encoder extracts the BEV embedding from surrounding camera images. For the textual branch, the text embedding is enriched by incorporating the knowledge graph embedding and then fed into a language encoder to generate language embedding.
\textbf{(b) Feature Alignment.} First, a set of shared learnable embeddings are employed to bridge the gap between the two branches' features. Moreover, a caption generation auxiliary task further enhances the alignment. Then, the resulting features are aligned with the contrastive loss.}

\label{fig: main_pipline} 
\end{figure*}
\subsection{Preliminaries}
\label{subsec:problem}

\mypara{Problem definition.}
% of text-scene pair $\{(T_i, I_i)\}_{i=1}^N$, where $(T_i, I_i)$ represents the paired relationship between the text sentence $T_i$ and the scene image $I_i$.
In this study, we concentrate on the task of text-scene retrieval on the autonomous driving dataset
with paired text sentence $\mathcal{T}$ and a scene collection with images $\mathcal{I}$.
Given a textual query $q$, the retrieval model returns a ranked list of scene images $L$. Here, $L_{i}$ represents the $i$-th ranked image in the list. 
The objective is to retrieve as many relevant images as possible from the top-$k$ ranked scene images $\mathcal{L}_{k}$. 
% We focus solely on the top-$k$ retrieval results, considering that users typically review only the top search results.

\mypara{Overall architecture.}
As illustrated in \cref{fig: main_pipline}, for feature extraction, our \name{} adapt a BEV encoder to extract the visual BEV embedding $b_{i}$, and the textual embedding is enriched by incorporating the knowledge graph embeddings and then embedded as $t_{i}$ from a language encoder.
For feature alignment, we leverage a set of shared learnable embeddings to bridge the gap between the two branches' features. The resulting features are aligned with the contrastive loss based on cosine similarity. Moreover, we also employ caption generation as an auxiliary task to enhance the alignment further.
During retrieval, we extract embedding $t$ for the given query text and compute the cosine similarity with the BEV vector $B=\{b_1, b_2, \dots, b_n\}$ of all images in the candidate pool to identify the top-$k$ similar scene images.

\subsection{Feature Extraction}
\label{subsec:extract}

\subsubsection{BEV Encoder.}
% BEV space learning can provide an intuitive representation of the world and enable the representation of objects in BEV, which is the optimal space for text-scene retrieval.
% For BEV feature extraction, our BEV encoder extracts BEV embeddings $B$ from visual sequence inputs $\mathcal{I}$; It can be any visual BEV encoder~\cite{li2022bevformer, huang2021bevdet, liu2022petr}
% The BEV encoder plays a crucial role in our approach as it allows for a more intuitive representation of the world and facilitates the effective representation of objects in the BEV space, which is particularly beneficial for text-scene retrieval tasks.
% In our framework, the BEV encoder is responsible for extracting BEV embeddings ($B$) from the visual sequence inputs ($\mathcal{I}$). We adopt a versatile visual BEV encoder, such as BEVFormer~\cite{li2022bevformer}, BEVDet~\cite{huang2021bevdet}, or PETR~\cite{liu2022petr}, which have demonstrated promising performance in similar tasks. These encoders are designed to capture relevant visual features in the BEV domain, thus enhancing the retrieval process.

The utilization of BEV space learning in autonomous driving allows for an intuitive representation of the world, making it an optimal choice for text-scene retrieval tasks. 
To extract BEV features, we employ a dedicated BEV encoder for generating BEV embeddings $B$ from the visual sequence inputs $\mathcal{I}$. It is worth noting that a wide range of visual BEV encoders can be employed for this purpose, such as BEVFormer~\cite{li2022bevformer}, BEVDet~\cite{huang2021bevdet}. These encoders have demonstrated exceptional performance in similar tasks and are capable of capturing informative visual features in the BEV domain.

\subsubsection{Text Encoder.}
To extract comprehensive semantics information from textual input, we utilize recent popular large language models as our text encoder, such as Llama~\cite{touvron2023llama} or GPT-3~\cite{floridi2020gpt}, which generates language embeddings $T$ from the text sentence $\mathcal{T}$. As these models~\cite{brown2020language, chowdhery2022palm, chung2022scaling, liu2023visual} have demonstrated great generalization power and common sense reasoning ability, indicating their potential to understand the scenarios in the realm of autonomous driving.

\subsubsection{Knowledge Graph Prompting}
% 这一节太长了，提升的结果又很少
% \label{subsec:KG}
% \mypara{Overview.}
% Integrating knowledge graph embedding into text-to-scene retrieval systems can enhance the retrieval process by considering the associations between different elements in the scene
% In traffic scene text descriptions, semantic information often exhibits discrete characteristics. 
% Traffic scenes typically involve specific objects, actions, and attributes. 
% Consequentially, 
Scene text inputs are usually described using discrete categories. 
For example, when discussing a traffic scene, discrete categories are used to describe the types of vehicles present (e.g., cars, trucks), their colors (e.g., red, blue), their actions (e.g., stopping, turning), and the scene's attributes (e.g., traffic lights, pedestrian crossings). 
However, only discrete semantic information is insufficient for text-scene retrieval,
as it lacks contextual comprehension and fails to capture relationships and associations between elements.
To address this, we propose incorporating knowledge graph embedding that provides associative information as a complement to enhance overall scene understanding.
% Specifically, we utilize a Graph Neural Network (GNN) to train the knowledge graph embedding of autonomous driving. 
% Specifically, we utilize the popular Knowledge Graph Embedding (KGE) technique to learn the knowledge graph embeddings of autonomous driving. 
% Each node in the graph corresponds to a keyword relevant to autonomous driving, and the embeddings associated with these nodes capture the associative representation of autonomous driving keywords. 
% % Each node in the graph corresponds to a keyword relevant to autonomous driving, and the associated embeddings capture the associative representation of these keywords. 
% Subsequently, these keyword knowledge graph embeddings are concatenated with the text embedding, thereby expanding the semantic representation of the encoded text.

% \begin{wrapfigure}{r}{8.5cm} %H为当前位置，!htb为忽略美学标准，htbp为浮动图形
\begin{figure}
    
\centering %图片居中
% \vspace{-1.2cm}
\includegraphics[width=1\textwidth]{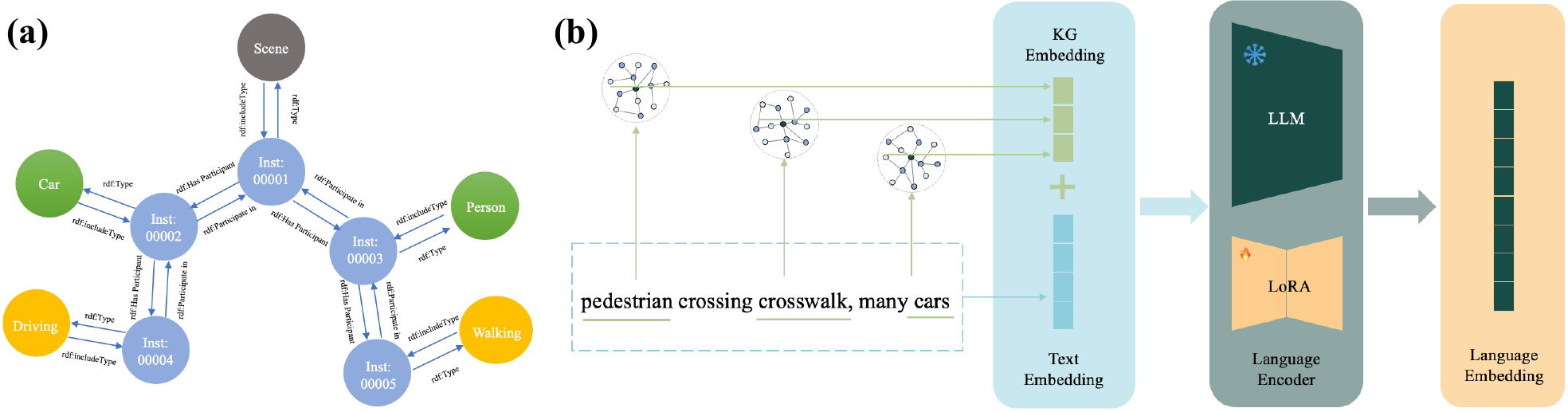}
 % \vspace{-0.3cm}
\caption{\textbf{Knowledge graph prompting.} \textbf{(a)} The knowledge graph embeddings are learned from the autonomous driving knowledge graph. 
Each node in the graph corresponds to a keyword relevant to autonomous driving, and the embeddings associated with these nodes capture the associative representation of autonomous driving keywords.  
\textbf{(b)} Subsequently, these keyword knowledge graph embeddings are concatenated with the text embedding, thereby expanding the semantic representation of the encoded text, and then embedded from a language encoder (zoom-in for better views).}
\label{fig: kg} 
% \vspace{-0.5cm}
% \end{wrapfigure}
\end{figure}

\mypara{Knowledge graph.}
% 有Graph的图可以画出来么 (替换table1)
% 学习到的feature最好也能可视化下
% \Tao{add KGs figure; add KGE feature vis}
Knowledge Graphs (KGs) are a type of multi-relational graphs that store factual knowledge in the real world. 
% Nodes in KGs represent real-world entities (e.g., names, events and products) and edges represent the relationships between entities. Normally, a KG can be efficiently stored as knowledge triples, where each triple consists of two entities and one factual relation between them (i.e., <head entity, relation, tail entity>). 
It is typically represented as $\mathcal{G} = \{\mathcal{E}, \mathcal{R}, \mathcal{S}\}$, where $\mathcal{E}$ denotes the set of entities (nodes), $\mathcal{R}$ denotes the set of relations (the types of edges) and $\mathcal{S}$ represents the relational facts (edges). 
Facts observed in $\mathcal{G}$ are stored as a collection of triples: $\mathcal{G} =\{ (h, r, t) \}$, where each triple consists of a head entity $h \in \mathcal{E}$, a tail entity $t \in \mathcal{E}$, and a relation $r \in \mathcal{R}$ between them, e.g., \textit{<scene, includes, car}> as illustrated in \cref{fig: kg} (a).
% % We use lowercase and bold characters to denote the embeddings of entities and relations. Specifically, for a fact triple $(h, r, t)$, we represent the embeddings of the head entity, tail entity, and the relation between them as $\mathbf{h}$, $\mathbf{t}$ and $\mathbf{r}$, respectively. 
% To acquire comprehensive knowledge of autonomous driving, it is essential to extract and generalize relationships within a KG across a wide range of instances. 
% % Constructing perceptual data into graph instances that can be effectively learned enables this process. To achieve this, 
% ADKG~\cite{nag2021towards} proposed a KG specifically tailored for autonomous driving, which is constructed using scene-aware data obtained from PandaSet~\cite{xiao2021pandaset}, and abstracts triplets to establish associations between perceptual instances, labels, and actions. \Tao{KG explain}
% Practically, we categorize the entities within the knowledge graph into three main groups: instances, objects, and movements.
% % 讲清楚instances, labels, and actions又对应什么， 自己分类的instances, objects, and movements又是什么
% % Within the knowledge graph, a predominant portion of nodes corresponds to instances, while object and movement nodes connect to numerous distinct instance nodes. 

\mypara{Knowledge graph embedding.}
% The goal of learning embeddings from a KG is to represent the entities and relations in low-dimensional vector space while also maintaining the semantics contained in the KG.
% The essential idea of KGE is to learn to embed entities and relations of a KG into a low-dimensional space (i.e. vectorial embeddings), where the embeddings are required to preserve the semantic meaning and relational structure of the original KG.
With the large-scale and complicated graph structure of the autonomous driving KG. We then need to learn knowledge graph embeddings (KGEs) to represent the entities and relations in low-dimensional vector space while also maintaining the semantics contained in the original KG.
As suggested by Wickramarachchi et al.~\cite{wickramarachchi2020evaluation}, we adopt the semantic transitional distance-based modeling, like TransE~\cite{bordes2013translating}, ConvE ~\cite{dettmers2018convolutional} and DistMult~\cite{yang2014embedding}
% , rather than semantic matching-based models. 
Specifically, we adopt a distance-based scoring function 
% to assess relationships within the graph. 
to optimize the embeddings.
For each triplet $(h, r, t)$, the scoring function is defined as follows:
\begin{equation}
    f_r(h,t) = -\Vert \mathbf{h} + \mathbf{r} -\mathbf{t} \Vert_{p},
\end{equation}
where $p = 1$ or $p = 2$, and $\mathbf{h}$, $\mathbf{t}$, and $\mathbf{r}$ represent the embedding of the head entity, tail entity, and the relation between entities, respectively. 
% The norm $\Vert \cdot \Vert_{norm}$ is defined as the $L1$ or $L2$ norm. 
With $\mathbf{r}$ represents a translation vector from $\mathbf{h}$ to $\mathbf{t}$, $\mathbf{h} + \mathbf{r} \approx \mathbf{t}$ when the triple ($h, r, t$) holds true.
% The scoring function of $f_r$ is defined as the negative distance between $\mathbf{h} + \mathbf{r}$ and $\mathbf{t}$, which is minimized with a margin-based hinge ranking loss function over the training process.
% and when the ($h, r, t$) holds, $f_r$ is expected to be large. 
With the scoring function, we learn KGEs that capture the relationships depicted by the triplets within the autonomous driving KG. 
% network: 注意，这并不是一个神经网络的模型，原文第3章提到了Loss更新的参数，是所有entities和relations的Embedding数据，每一次SGD更新的参数就是一个Batch中所有embedding的值。也就是说，是通过训练，逐步的调整entities和relations的嵌入表示的数据，最后训练停止，输出所有embedding的结果
% \Tao{network}

\mypara{Semantic representation fusion.}
Then, we utilize the obtained knowledge graph embeddings to enrich the semantic representation of the original text embedding. Practically, as illustrated in \cref{fig: kg} (b),  we index the keywords from the graph that appear in the text input and concatenate their knowledge graph embeddings into the text embedding sequence in the order of their occurrence.
% By incorporating these keywords' knowledge graph embeddings into the text embedding, we enrich the overall semantic representation of the text.

\subsection{Feature Alignment}
\label{subsec:alignment}
\subsubsection{Shared Cross-modal Embedding}
\begin{wrapfigure}{r}{5.5cm} %H为当前位置，!htb为忽略美学标准，htbp为浮动图形
% \begin{figure}
\vspace{-24pt}
\centering %图片居中
% \vspace{-1.2cm}
\includegraphics[width=0.35\textwidth]{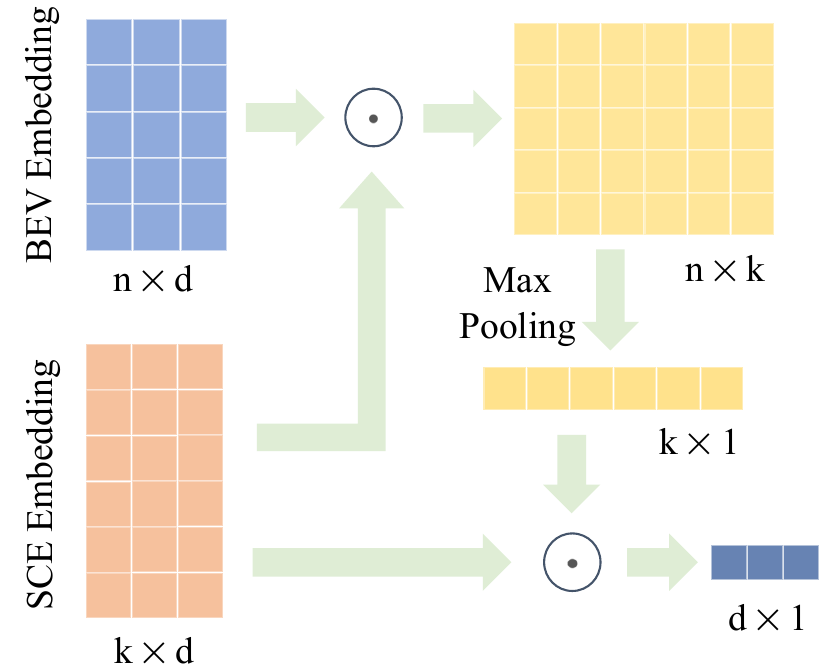} %插入图片，[]中设置图片大小，{}中是图片文件名
 % \vspace{-0.3cm}
\caption{\textbf{Detail architecture of SCE.}} %最终文档中希望显示的图片标题
\label{fig: scp} %用于文内引用的标签
\vspace{-15pt}
\end{wrapfigure}
% \end{figure}
% \label{subsec:SCE}
% In this section, we explain how the cross-modal interaction in our method is performed between the BEV and the text branch. 
% We realised that the components in both BEV and text branches are pre-trained on single-modal datasets. Meanwhile, most of the parameters of these components have to remain frozen to maintain an affordable training cost. 
% Our objective is to utilize these prompts to map the Bird's Eye View (BEV) features and textual features onto the same manifold space. 
% This facilitates the alignment of divergent modal information found in both branches.
Through the BEV encoder and LLM text encoder with KGP, we obtain the BEV feature and language embedding in two different feature spaces.
% , resulting in substantial dimensional differences between them.
Then directly aligning these cross-modal features leads to unsatisfactory outcomes.
% To address this challenge, inspired by the Q-Former in BLIP2~\cite{li2023blip}, we propose Shared Cross-modal Embedding (SCE) with a set of shared learnable embeddings to bridge this gap between the two modalities by re-projecting their respective features into a shared feature space, similar to an attention mechanism.
To address this challenge, we propose the Shared Cross-modal Embedding (SCE) approach, inspired by the Q-Former in BLIP2~\cite{li2023blip}. SCE utilizes a set of shared learnable embeddings to bridge the gap between the two modalities by re-projecting their respective features into a shared feature space, similar to an attention mechanism.

Specifically, as illustrated in \cref{fig: scp}, the learnable shared cross-modal embeddings are denoted as $C=\{c_1, c_2, \dots, c_k\}$, and the BEV features can be reshaped and compressed into a sequence of BEV embeddings as $B=\{b_1, b_2, \dots, b_n\}$.
% For each embedding $t_i$ from $T$ in the shared cross-modal embeddings sequence, we compute the similarity between each feature $b_j$ in the BEV sequence and the token $t_i$ as $s_{ij} = \text{sim}(t_i, b_j)$, where $\text{sim}$ represents cosine similarity. We then obtain the maximum similarity value between the BEV embeddings sequence $B$ and the token $c_i$ as follows:
For each embedding $c_i$ from $C$ in the shared cross-modal embeddings sequence, the similarity between each feature $b_j$ in the BEV sequence and the embedding $c_i$ is computed as $s_{ij} = sim (c_i, b_j)$, where $sim$ represents cosine similarity. Then the maximum similarity is obtained between the BEV embeddings sequence $B$ and the embedding $t_i$ as $r_i = \max_j(s_{ij})$.
% \begin{equation}
% r_i = \mathop{max}\limits_{j}(s_{ij}).
% \end{equation}
% \begin{equation}
% r_i = \max_j(s_{ij}).
% \end{equation}
For cross-modal embeddings $C$, we have $R = \{r_1, r_2, \dots, r_k\}$. Then, with the softmax function, $R$ can be converted into weights: 
% \begin{equation}
$
    w_{i}^{b}=\frac{e^{r_{i}}}{\sum e^{r}}.
$
% \end{equation}
The re-projected BEV embeddings are obtained by multiplying the weights with the shared cross-modal embeddings:
$
B^{\prime} = \{b_1^{\prime}, b_2^{\prime}, \dots, b_n^{\prime}\} = \{w_1^{b} c_1, w_2^{b} c_2, \dots, w_k^{b} c_k\}
$.
Similarly, for the language embeddings $T$, we obtain $T^{\prime}=\{w_1^{t} c_1, w_2^{t} c_2, \dots, w_k^{t} c_k\}$. 
% Similarly, for the language embeddings $T=\{t_1, t_2, \dots, t_k\}$, we obtain $T^{\prime}=\{w_1^{t} c_1, w_2^{t} c_2, \dots, w_k^{t} c_k\}$. 
Subsequently, the obtained BEV and language embeddings $B^{\prime}$ and $L^{\prime}$ are aligned with contrastive loss:
% \begin{equation}
%     L_q= -log\frac{exp(q \cdot k_+/\tau)}{\sum\limits_{i=0}^kexp(q\cdot k_i/\tau)}
%     \label{eq:losssce}
% \end{equation}
\begin{equation}
     \mathcal{L}_{SCE}  = \mathcal L_{\rm cl}^{\rm t2s}  +  \mathcal L_{\rm cl}^{\rm s2t},
     \label{eq:losssce}
\end{equation}
\vspace{-6pt}
\begin{small}
\begin{equation}
    \mathcal L_{\rm cl}^{\rm t2s}  =\frac 1 N \sum_{i=1}^N-\log (\frac{\exp(sim({t}_i^{\prime}, b_i^{\prime})/\tau)}{\sum_{j=1}^N \exp(sim({t}_i^{\prime}, b_j^{\prime})/\tau)}), 
    \mathcal L_{\rm cl}^{\rm s2t}  =\frac 1 N \sum_{i=1}^N-\log (\frac{\exp(sim(b_i^{\prime}, {t}_i^{\prime})/\tau)}{\sum_{j=1}^N \exp(sim({b}_i^{\prime}, t_j^{\prime})/\tau)}),
\end{equation}
\end{small}

where $N$ is the batch size and $\tau$ is a temperature hyper-parameter, and $t2s$ and $s2t$ denotes text-to-scene and scene-to-test respectively.

Practically, the cross-modal embeddings shared between the BEV and text branches serve as a bridge in a shared feature space. 
% It enables the preservation of the original feature shapes while aligning the two modalities within the same embedding space. 
% By sharing the embeddings, we maintain the distinct shapes of the original features while achieving alignment between the BEV and text modalities within a unified embedding space.
It enables maintaining the distinct shapes of the original features while achieving alignment between the BEV and text modalities within a unified embedding space.

\subsubsection{Caption Generation}
% \label{subsec:caption}
% Additionally, 为了进一步加强bev embedding 和language embedding的对齐， we introduce a caption generation task based on the BEV embeddings as an auxiliary component for model training. We utilize a lightweight transformer-based decoder, while the corresponding text description to the BEV sample serves as the supervision label for this task.
To further enhance the alignment between the BEV embedding and the language embedding, we introduce an auxiliary caption generation task based on the BEV embeddings.
%as an auxiliary component during model training.
We utilize a lightweight transformer-based decoder, with the corresponding text description of the BEV sample serving as the supervision label:
\begin{equation} 
    \mathcal{L}_{CG} = \text{CrossEntropy}(P_{logits},T), 
    \label{eq:losscg}
\end{equation}
where $P_{logits}$ is the logits of  predictive text tokens and $T$ is the target text tokens.
By incorporating this caption generation task, we strengthen the relationship between the BEV embedding and the generated textual description, thereby improving the overall text-scene retrieval performance.
% \paragraph{Caption Generation} We introduce a MLLM $g_{\rm mllm}(\cdot,\cdot)$ to generate a corresponding caption for each image in the dataset. Specifically, we design a prompt template $P_{\rm{cap}}(type, k)$ to guide the MLLM to obtain a brief caption for each image under constrained conditions, where $type$ and $k$ are two dataset-specific parameters 
% to simulate the type and length of modified text in the real dataset. For an image $I_i$ in the candidate image set, we input $I_i$ and $P_{\rm{cap}}$ together into the MLLM to obtain the corresponding caption $C_i$:
% \begin{equation}
%     C_i=g_{\rm mllm}(I_i,P_{\rm{cap}}(type,k)).
% \end{equation}\label{equ:imgcap}Then we can obtain M image-text pairs $\{(I_1,C_1),...,(I_M,C_M)\}$. In practice, the $P_{\rm{cap}}$ used in this work is written as follows:
% % \begin{tcolorbox}
% % \texttt{<{\color{blue} image tokens}> Please briefly describe the \{{\color{blue} type}\} in \{{\color{blue} k}\} words.}
% % \end{tcolorbox}
% \begin{tcolorbox}\small
% \texttt{Please briefly describe the \{{\color{blue} type}\} in \{{\color{blue} k}\} words.}
% \end{tcolorbox}

\subsection{Overall Optimization}

% loss写清楚，公式写清楚
To summarize, the overall optimization target of our \name{} is formulated as:
\begin{equation}
\mathcal{L} = \mathcal{L}_{SCE}  + \lambda  \mathcal{L}_{CG} 
\end{equation}
where $\mathcal{L}_{SCE}$ and $\mathcal{L}_{CG}$ are defined in \cref{eq:losssce} and \cref{eq:losscg}, and $\lambda$ indicate the weight balance coefficient.
% and set to 0.15 by default.

\section{\nusr{} Dataset} 
\label{sec:nusr}
% train_image_to_desc.json : 848 / 34149
% train_image_to_desc_new.json : 10350 / 34149
% merged_nuscene_qa_train.json : 29800 / 34149

% \begin{wrapfigure}{r}{8.5cm} %H为当前位置，!htb为忽略美学标准，htbp为浮动图形
\begin{figure}[h!]
\centering %图片居中
\vspace{-6pt}
\includegraphics[width=1\textwidth]{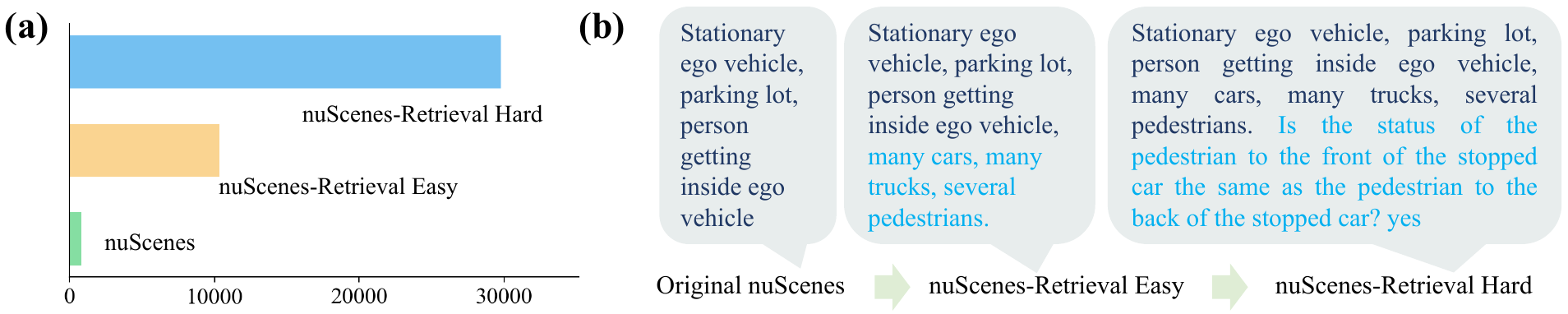} %插入图片，[]中设置图片大小，{}中是图片文件名
 % \vspace{-0.3cm}
 \vspace{-16pt}
\caption{\textbf{(a)} The number of text descriptions. \textbf{(b)} A case on the \nusr{} dataset.}
\label{fig: dataset} %用于文内引用的标签
\vspace{-9pt}
% \end{wrapfigure}
\end{figure}

Although we have developed the \name{} framework for autonomous driving retrieval, the driving community lacks well-structured retrieval datasets for effective evaluation. The only available dataset with scene sensor data and open text descriptions is the nuScenes dataset~\cite{caesar2020nuscenes}. 
% However, the provided text descriptions in nuScenes are very simplistic and lack detailed scene information, resulting in a significant amount of repetition in the original text descriptions due to the lack of detailed information.
% As shown in \cref{fig: dataset} (a), the nuScenes dataset consists of over 30,000 samples but only has 848 distinct text sentences available, indicating a relatively low diversity in the textual descriptions.
However, the provided text descriptions in nuScenes are very simplistic and lack detailed scene information, as shown in \cref{fig: dataset} (a), the nuScenes dataset consists of over 30,000 samples but only has 848 distinct text sentences available, which leads to significant repetition in the provided text descriptions.
% resulting in a significant amount of repetition in the original text descriptions due to the lack of detailed information.
% , indicating a relatively low diversity in the textual descriptions.
% The nuScenes dataset, as depicted in Figure (a), includes more than 30,000 samples; however, it suffers from a limited variety of text sentences, with only 848 distinct descriptions. This lack of detailed information leads to significant repetition in the provided text descriptions.
To address these limitations, we have further constructed the \nusr{} dataset based on the nuScenes dataset, and the toolkit codes are attached in the supplement materials and will be public.

\mypara{\nusr{} Easy.} 
As shown in \cref{fig: dataset} (b), to reduce text repetition, we enhance the original captions by supplementing them with perception results. Specifically, for each keyframe sample, we extract obstacle information from the detection labels. Moreover, we quantify the frequency of occurrence for each obstacle type, then render the obstacles with quantity descriptors, such as "many cars, several trucks, one bus". The supplemented text is concatenated after the original caption text, resulting in a more comprehensive scene description for training and evaluation. This modified dataset is referred to as \textit{\nusr{} Easy}, which provides more than 10k text descriptions.
% 2-5 several；5 many；1 one

\vspace{-3pt}
\mypara{\nusr{} Hard.} 
To further improve the informative diversity of the textual descriptions, we have noticed research efforts, i.e., NuScenes-QA~\cite{qian2023nuscenes}, which is a visual question-answer (VQA) dataset generated from the nuScenes dataset, consisting of over 34,000 scenes with more than 460,000 question-answer pairs. These pairs are generated from perception information using manually designed QA templates and cover various aspects, including object detection, scene classification, ego-vehicle decision-making, and decision reasoning.
As shown in \cref{fig: dataset} (b), to further enrich the text input with semantic information related to potential decision descriptions, we concatenate the questions and answers together and append them at the end of the previous description. This extended dataset is dubbed \textit{\nusr{} Hard} with 29k text descriptions.
% To further improve the informative diversity of the textual descriptions, we have noticed research efforts, i.e., NuScenes-QA~\cite{caesar2020nuscenes}, aimed at expanding the textual and semantic information of multi-view autonomous driving data. 
% NuScenes-QA is a visual question-answer (VQA) dataset generated from the nuScenes dataset, consisting of over 34,000 scenes with more than 460,000 question-answer pairs. 
% These pairs are generated from perception information using manually designed QA templates and cover various aspects, including object detection, scene classification, ego-vehicle decision-making, and decision reasoning. 
% As shown in \cref{fig: dataset} (b), to further enrich the text input with semantic information related to potential decision descriptions, we concatenate the questions and answers together and append them at the end of the original nuScenes description. This extended dataset is dubbed \textit{\nusr{} Hard}.
\section{Experiment}
\vspace{-5pt}
In this section, we first introduce the experimental setup in \cref{subsec: setup}. Then, we compare \name{} with previous approaches in \cref{subsec:mainres}. The analysis of each component is presented in \cref{subsec:ablation}. In the end, we provide qualitative results in \cref{subsec: vis}. 

\subsection{Experimental Setup}
\label{subsec: setup}
\vspace{-5pt}
\mypara{Datasets and Metric.}
For the retrieval dataset, we adopt the \nusr{} in \cref{sec:nusr} based on the nuScenes dataset~\cite{caesar2020nuscenes}.
We use (R@K, K=1,5,10) as our evaluation metrics of recall accuracy, which is the most commonly used evaluation metric in the retrieval tasks and is the abbreviation for recall at $k$-th in the ranking list, defined as the proportion of correct matchings in top-$k$ retrieved results. 
The implementation details are provided in the supplementary material.

\subsection{Main Results}
\label{subsec:mainres}
\begin{table*}[t]
  \centering 
  \small
  \caption{\textbf{Comparisons with different retrieval methods.}
  % Our \name{} achieves remarkable retrieval performance.
  }
   \addtolength{\tabcolsep}{0.8pt}
    \begin{tabularx}{\linewidth}{l|c|ccc|ccc}
\toprule

\multirow{2}{*}{Method}  &  \multirow{2}{*}{Retrieve Space} & \multicolumn{3}{c|}{Text Retrieval} & \multicolumn{3}{c}{Scene Retireval}  \\
      \cmidrule(r){3-8} &   & R@1 & R@5 & R@10  & R@1 & R@5 & R@10 \\
       
   \midrule
        \multicolumn{3}{l}{\textit{\textbf{\nusr{} Easy}}}    \\
        \midrule
        \multirow{2}*{SigLIP-Base~\cite{zhai2023sigmoid}}
        &Front View & 0.3683          & 0.7640          & 0.8613          & 0.3924          & 0.7863          & 0.8698       \\
     & Surrounding View & 0.3433          & 0.7625          & 0.8593          & 0.3597          & 0.7740          & 0.8672 \\
    %  & BEV Embedding \\
   %      \midrule
   %      \multirow{2}*{CLIP-ConvNeXT-Base~\cite{radford2021learning}} 
   % &Front View Image & 0.4878          & 0.7435          & 0.8267          & 0.5737          & 0.8320          & 0.8901       \\
   %   & Surrounding View Image & 0.4169          & 0.6617          & 0.7449          & 0.5157          & 0.7687          & 0.8294 \\
   %  %  & BEV Embedding \\
       \midrule
        \multirow{2}*{CLIP-ViT-Base~\cite{radford2021learning}} 
   &Front View & 0.4377          & 0.8610          & 0.9569          & 0.4421          & 0.9003          & 0.9795          \\
  & Surrounding View & 0.4846          & 0.9085          & 0.9815          & 0.4644          & 0.9258          & 0.9845          \\
 % & BEV Embedding      & 0.7875          & 0.9757          & 0.9909          & 0.8194          & 0.9812          & 0.9906          \\
  \midrule
   \multirow{2}*{EVA02-Base~\cite{fang2023eva}} 
   &Front View & 0.4919          & 0.7306          & 0.7977          & 0.5585          & 0.7807          & 0.8440          \\
     & Surrounding View & 0.4369          & 0.7153          & 0.7986          & 0.5181          & 0.7896          & 0.8637 \\
    %  & BEV Embedding \\
       \midrule

 \rowcolor{mygray} \bf \name{} (Ours) & BEV Space & \bf 0.8578          & \textbf{0.9954} & \textbf{0.9994} & \textbf{0.8766} & \textbf{0.9971} & \textbf{0.9997} \\

   \midrule
     \midrule
        \multicolumn{3}{l}{\textit{\textbf{\nusr{} Hard}}}    \\
        \midrule
        \multirow{2}*{SigLIP-Base~\cite{zhai2023sigmoid}} 
   &Front View & 0.2687          & 0.6573         & 0.7661          & 0.2850          & 0.6691          & 0.7670       \\
     & Surrounding View & 0.2594          & 0.6487          & 0.7379          & 0.2843          & 0.6501          & 0.7365 \\
     % & BEV Embedding \\
   %      \midrule
   %      \multirow{2}*{CLIP-ConvNeXT-Base~\cite{radford2021learning}} 
   % &Front View Image & 0.          & 0.          & 0.          & 0.          & 0.          & 0.       \\
   %   & Surrounding View Image & 0.          & 0.          & 0.          & 0.          & 0.          & 0. \\
   %  %  & BEV Embedding \\
       \midrule
          \multirow{2}*{CLIP-ViT-Base~\cite{radford2021learning}} 
   &Front View & 0.2829          & 0.6501          & 0.7886          & 0.2986          & 0.6888          & 0.7863      \\
     & Surrounding View & 0.2904          & 0.6619          & 0.7953          & 0.3163          & 0.7066          & 0.7896      \\
     % & BEV Embedding \\
      \midrule
   \multirow{2}*{EVA02-Base~\cite{fang2023eva}} 
   &Front View & 0.2908          & 0.6763          & 0.7860          & 0.3064          & 0.6980          & 0.7936          \\
     & Surrounding View & 0.2774          & 0.6395          & 0.7318          & 0.2965          & 0.6538          & 0.7430 \\
       % & BEV Embedding \\
       \midrule

 \rowcolor{mygray} \bf \name{} (Ours) & BEV Space & \bf 0.6608          & \textbf{0.9912} & \textbf{0.9997} & \textbf{0.6790} & \textbf{0.9862} & \textbf{0.9991} \\

\bottomrule
\end{tabularx}
\vspace{-8pt}
\label{tab: main_results}
\end{table*}
% Firstly we used only the front view camera to train and evaluate, and then we attempted to concatenate 6 view images into one single image. 
In \cref{tab: main_results}, we compare our \name{} with state-of-the-art retrieval methods, i.e., CLIP-ViT-Base~\cite{radford2021learning}, SigLIP-Base~\cite{zhai2023sigmoid}, and EVA02-Base~\cite{fang2023eva} on the multi-level \nusr{} dataset.
To ensure a fair comparison, we extended the previous methods to include six surrounding images. 
From the results, it can be observed that our \name{} achieves impressive performance (85.78\% and 87.66\% top-1 accuracy) and outperforms all the other methods by a large margin, both from the front view and six surrounding views. This indicates that our method is effective in retrieving scenes in the BEV space and demonstrates a significant capability to understand the global context and retrieve complex traffic scenarios.
Furthermore, across different levels of the \nusr{} dataset, our method consistently achieves superior results. Particularly, a more pronounced improvement is achieved in the Hard level, indicating that our \name{} framework with well-designed modules, is capable of better understanding and retrieving more complex autonomous driving scenes.

\subsection{Ablation study}
\label{subsec:ablation}
% In this section, we verify the effect of each of our proposed methods on the retrieval results and validate the effectiveness of the methods through multiple sets of ablation experiments. We discuss the effect of the large language model, knowledge graph, Shared Cross-Modal Prompt, and caption generation tasks respectively. 

\begin{table*}[t]
  \centering 
  \caption{\textbf{Ablations on the proposed modules of \name{}.}}
   \addtolength{\tabcolsep}{2pt}
    \begin{tabularx}{\linewidth}{cccc|ccc|ccc}
\toprule
   % SCP & KG & CG          
   
   % & B2T\_R1        & B2T\_R5        & B2T\_R10       & T2B\_R1         & T2B\_R5         & T2B\_R10        \\ 
  \multirow{2}{*}{BEV} & \multirow{2}{*}{SCE}  &  \multirow{2}{*}{KGP} & \multirow{2}{*}{CG} & \multicolumn{3}{c|}{Text Retrieval} & \multicolumn{3}{c}{Scene Retireval}  \\
      \cmidrule(r){5-10} & & & & R@1 & R@5 & R@10  & R@1 & R@5 & R@10 \\
      
\midrule
  
  \xmarkg & \xmarkg  &  \xmarkg &  \xmarkg & 0.4846          & 0.9085          & 0.9815          & 0.4644          & 0.9258          & 0.9845          \\
 
  \colorbox{mycyan}{\cmark} &    \xmarkg & \xmarkg  &  \xmarkg                                 & 0.7875          & 0.9757          & 0.9909          & 0.8194          & 0.9812          & 0.9906          \\
 \cmarkg & \colorbox{mycyan}{\cmark}  &  \xmarkg  &  \xmarkg   & 0.8352          & 0.9944 & 0.9988          & 0.8431 & 0.9962          & 0.9991 \\
 % \xmarkg &  \colorbox{mycyan}{\cmark}  &  \xmarkg      & 0.8059          & 0.9783          & 0.9947          & 0.8584          & 0.9909          & 0.9959          \\
% \cmarkg &  \cmarkg &\colorbox{mycyan}{\cmark}  &  \xmarkg       & \textbf{0.8599} & 0.9947          & 0.9994          & 0.8757          & 0.9968          & 0.9994          \\
\cmarkg &  \cmarkg &\colorbox{mycyan}{\cmark}  &  \xmarkg       & 0.8469 & 0.9947          & 0.9994          & 0.8557          & 0.9968          & 0.9994          \\
  \rowcolor{mygray} \cmarkg & \cmarkg & \cmarkg  & \colorbox{mycyan}{\cmark}  & \bf 0.8578          & \textbf{0.9954} & \textbf{0.9994} & \textbf{0.8766} & \textbf{0.9971} & \textbf{0.9997} \\ 
   % \cmarkg  & \xmarkg  &  \colorbox{mycyan}{\cmark} & \textbf{0.8573} & 0.9930          & \textbf{0.9988} & \textbf{0.8751} & \textbf{0.9974} & \textbf{0.9991} \\
\bottomrule
\end{tabularx}
\vspace{-8pt}
\label{tab: ablation_nus}
\end{table*}

\mypara{Ablations on the proposed modules of \name{}.}
In \cref{tab: ablation_nus}, we validate our proposed modules for model’s performance on \nusr{}.
% In \cref{tab: ablation_nus}, we validate our proposed modules on the \nusr{} Easy level.
It is clear that incorporating each module facilitates the understanding of complex traffic scenarios and leads to performance gain in retrieving.
Specifically, the BEV space learning (BEV) significantly improves the baseline of 35.5\% R@1 scene retrieval, which reflects the BEV providing an intuitive representation of the world, which is most desirable for text-scene retrieval.
Then the Shared Cross-modal Embedding (SCE) achieves remarkable improvements of 5.73\% R@1 scene retrieval by aligning the BEV and text modalities within a unified embedding space. It's noted that when without the SCE module, we utilize an MLP layer for feature mapping.
The Knowledge Graph Prompting (KGP) and Caption Generation (CG) modules also obtained consist boosts.
% with 0.26\% and 0.09\% R@1 respectively.
Moreover, combining these modules leads to further improvements, indicating that they effectively contribute to retrieving complex autonomous driving scenes.

\begin{table*}[t]
  \centering 
  \caption{\textbf{Ablations on each component of \name{}.}}
   \addtolength{\tabcolsep}{4pt}
    \begin{tabularx}{\linewidth}{c|ccc|ccc}
\toprule 

   % Module                     & B2T\_R1        & B2T\_R5        & B2T\_R10       & T2B\_R1         & T2B\_R5         & T2B\_R10        \\
   \multirow{2}{*}{Module}  & \multicolumn{3}{c|}{Text Retrieval} & \multicolumn{3}{c}{Scene Retireval}  \\
      \cmidrule(r){2-7}   & R@1 & R@5 & R@10  & R@1 & R@5 & R@10 \\

      \midrule 
  \multicolumn{3}{l}{\textit{\textbf{
BEV Encoder}} }   \\
    \midrule
BEVDet~\cite{huang2021bevdet} & 0.7955          & 0.9929 & 0.9966 & 0.8021 & 0.9917 & 0.9986 \\ 
  % PETR~\cite{liu2022petr} \\ 
  \rowcolor{mygray} BEVFormer~\cite{li2022bevformer}  & \bf 0.8578          & \textbf{0.9954} & \textbf{0.9994} & \textbf{0.8766} & \textbf{0.9971} & \textbf{0.9997} \\

          \midrule 
  \multicolumn{3}{l}{\textit{\textbf{
Text Encoder}}}    \\
    \midrule
     BERT~\cite{devlin2018bert}     &  0.6409          & 0.9129          & 0.9557          & 0.5594          & 0.8915          & 0.9384          \\ 
Llama2~\cite{touvron2023llama}      & 0.7244          & 0.9472          & 0.9713          & 0.7030          & 0.9507          & 0.9730          \\
 \rowcolor{mygray} Llama2 w/ LoRA~\cite{hu2021lora} & \textbf{0.7875} & \textbf{0.9757} & \textbf{0.9909}          & \textbf{0.8194} & \textbf{0.9812} & \textbf{0.9906} \\

      \midrule
        \multicolumn{3}{l}{\textit{\textbf{
Knowledge Graph Prompting}  }}  \\
    \midrule
   % --      &           &         &           &           &           &           \\
   TransE~\cite{bordes2013translating} & 0.8487          & 0.9914          & 0.9994          & 0.8559          & 0.9914          & 0.9994          \\

     ConvE~\cite{dettmers2018convolutional}    & 0.8501 & 0.9928 & 0.9994          & 0.8602 & 0.9956 & 0.9994 \\ 
      \rowcolor{mygray}  Distmult~\cite{yang2014embedding} & \textbf{0.8578} & \textbf{0.9954} & \textbf{0.9994}          & \textbf{0.8766} & \textbf{0.9971} & \textbf{0.9997}          \\ 
      \midrule
              \multicolumn{3}{l}{\textit{\textbf{
Shared Cross-modal Embedding}  }}  \\
    \midrule
   $k$ =    1024   & 0.8487 & 0.9912 & 0.9994          & 0.8602 & 0.9971 & 0.9991        \\
     $k$ =  2048  & 0.8473 & 0.9937 & 0.9997          & 0.8689 & 0.9968 & 0.9994 \\ 
      \rowcolor{mygray}  $k$ = 4096 & \textbf{0.8578} & \textbf{0.9954} &  \bf 0.9994          & \textbf{0.8766} & \textbf{0.9971} &  \bf 0.9997          \\

\bottomrule 
\end{tabularx}
\vspace{-4pt}
\label{tab: ablation_all}
\end{table*}
% baseline + KG
\mypara{Ablations on the BEV Encoder.}
In the first block of \cref{tab: ablation_all}, we investigate different BEV encoders for converting 2D features into 3D space, including BEVDet~\cite{huang2021bevdet} and BEVformer~\cite{li2022bevformer}. 
% Our approach achieves excellent retrieval results with both decoders, yielding comparable performance, which demonstrates the strong generalization ability of the proposed approach.
% Consistent improvements ranging from 5.2\% to 6.3\% R@1 scene retrieval can be observed across different BEV encoders, which demonstrates the strong generalization ability of the proposed approach.
%  Our approach achieves excellent results with both decoders, yielding comparable performance. The results obtained with DETR3D are slightly higher, leading us to select DETR3D as our default decoder.
Our approach achieves favorable retrieval results with both decoders, which demonstrates our generalization ability. While BEVFormer introduces sequential temporal modeling with spatiotemporal transformers which efficiently extract scene information into BEV space, leading to optimal results.

\mypara{Ablations on the Text Encoder.} 
In the second block of \cref{tab: ablation_all}, we conducted experiments to validate different text decoders for our proposed method, i.e., BERT~\cite{devlin2018bert} and Llama2~\cite{touvron2023llama}. Additionally, we also utilize LoRA~\cite{hu2021lora} to fine-tune the Llama2 model.
The results show that the Llama2 decoder is more effective than BERT in encoding relevant and accurate scene descriptions.
Furthermore, fine-tuning the Llama2 models using the LoRA leads to gains of 11.64\% in the R@1 scene retrieval. This improvement can be attributed to the fact that the pre-training task of Llama2 contains fewer traffic scenarios, fine-tuning tailors the models to the specific context of autonomous driving scenes.
% In the second block of \cref{tab: ablation_all}, we conducted experiments to validate different text decoders for our proposed method. Specifically, we tested two popular decoders: BERT~\cite{devlin2018bert} and Llama2~\cite{touvron2023llama}. Additionally, we also utilize LoRA~\cite{hu2021lora} to fine-tune the Llama2 model. 
% The results show that the Llama2 decoder significantly outperforms the BERT decoder in all retrieval metrics. This suggests that the Llama2 decoder is more effective in encoding relevant and accurate scene descriptions.
% Furthermore, fine-tuning the Llama2 models using the LoRA leads to significant gains of 11.64\% in the R@1 scene retrieval. This improvement can be attributed to the fact that the pre-training task of Llama2 contains fewer autonomous driving scenarios. Fine-tuning the Llama2 with the autonomous driving datasets can bridge this gap and better tailor the models to the specific context of autonomous driving scenes.

\mypara{Ablations on Knowledge Graph Prompting.}
In the third block of \cref{tab: ablation_all}, we valid different knowledge graph embedding techniques, including TransE~\cite{bordes2013translating}, ConvE~\cite{dettmers2018convolutional} and DistMult~\cite{yang2014embedding}.
% By incorporating learned KGEs from all three models, we observed improved comprehension of the textual elements and consistent improvements compared to the baseline method. 
% This suggests that leveraging KGEs enhances the ability of the model to understand and represent the relationships between entities and relations in the KG, leading to improved retrieval performance.
By incorporating learned KGEs from all three models, we observed consistent improvements. 
This suggests that leveraging KGEs enhances the ability to understand and represent the relationships between entities and relations in the original KG, leading to improved retrieval performance.
% Our approach achieves excellent results with both decoders, yielding comparable performance. The results obtained with DETR3D are slightly higher, leading us to select DETR3D as our default decoder.
% Among these techniques, DistMult achieved the optimal result, which effectively captures the semantics of knowledge graph triples of autonomous driving.
The results obtained with DistMult are slightly higher, leading us to select it as our default KGE extractor.

\mypara{Ablations on Shared Cross-modal Embedding.}
In the last block of \cref{tab: ablation_all}, we investigate the influence of the lantern shape of the shared cross-modal embeddings. The results reveal the larger lantern shape leads to better retrieval performance as they have the ability to adequately align two modalities' features. The shape of $k = 4096$ is sufficient to generate favorable retrieval results.

\subsection{Qualitative Results}
\label{subsec: vis}
\begin{figure}[ht]
        \centering
        \includegraphics[width=1\linewidth]{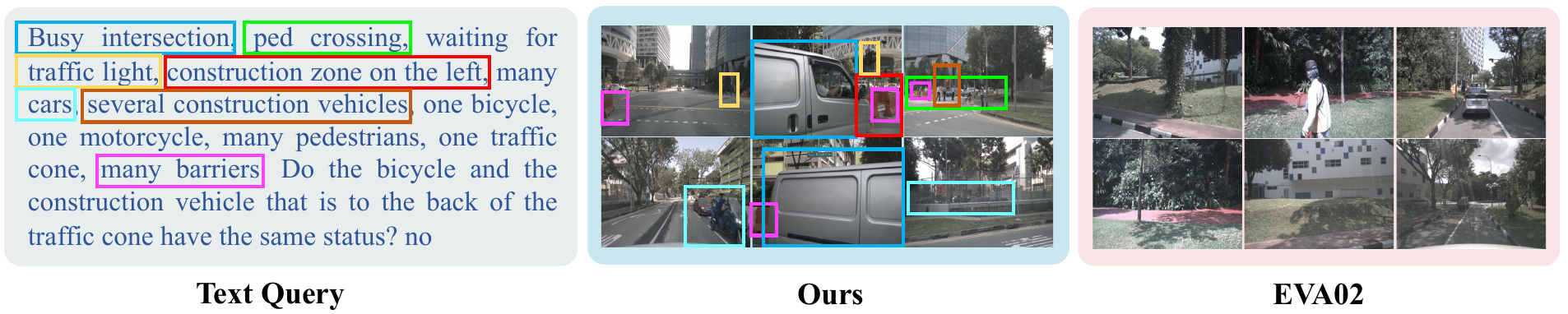}
        \caption{\textbf{R@1 scene retrieval results.} Our \name{} achieves more precise retrieval results.}
        \label{fig:vis}
        \vspace{-4pt}
\end{figure}
In \cref{fig:vis}, we present the qualitative results on \nusr{} dataset. We compare our \name{} with the previous state-of-the-art retrieval method, EVA02~\cite{fang2023eva}.
EVA02, as an image space retrieval model, fails to find scenes that correspond well to the textual queries. In contrast, our \name{} achieves more precise retrieval results, which retrieves scenes in BEV space and demonstrates a significant capability to understand the global context
\section{Conclusion}
In this paper, we propose the novel \name{} framework for text-scene retrieval in autonomous driving, which retrieves scenes in BEV space and demonstrates a significant capability to understand the global context and retrieve complex traffic scenarios.
For text sentence representation, we leverage an LLM and incorporate knowledge graph embeddings to comprehensively understand complex textual descriptions, offering a higher level of semantic richness in language embedding.
To align the features, we propose Shared Cross-modal Embedding, which utilizes a set of shared learnable embeddings to bridge the gap between the BEV features and language embeddings in different feature spaces. We also leverage a caption generation task to further enhance the alignment.
Moreover, we establish a multi-level retrieval dataset, \nusr{}, based on the nuScenes dataset, on which our \name{} achieves state-of-the-art performance with remarkable improvement, and extensive ablation experiments demonstrate the effectiveness of our proposed method.

% \mypara{Social Impact and Limitations.}
% Our framework represents the first text-scene retrieval approach in autonomous driving. 
% At present, we primarily focus on the image representation of scenes. However, our framework can be expanded to incorporate other sensors, such as LiDAR or a fusion of LiDAR and cameras, which can be achieved by merely substituting the corresponding BEV encoder\cite{bai2022transfusion, liang2022bevfusion}.
% % In our current implementation, 
% Currently, we employ pre-trained BEV and fine-tuned language encoders to effectively perform retrieval tasks. It would be valuable to explore the potential of leveraging our framework for pre-training tasks, similar to the pioneering CLIP~\cite{radford2021learning}.
% Additionally, in this study, we conducted experiments with various efficient learning methods for KGE. In the future, more sophisticated networks, e.g., Graph Neural Networks (GNN)~\cite{wu2020comprehensive}, could be introduced to enhance the learning of embeddings.
% We hope that our work can inspire future research in this field.

{
    \small
    \bibliographystyle{ieeenat_fullname}
    \bibliography{main}
}
\clearpage
\newpage
\appendix
\phantomsection
\addcontentsline{toc}{section}{Supplementary Materials}
\section*{\centering \name{} Supplementary Materials}

\appendix
\section{Social Impact and Limitations.}
Our framework represents the first text-scene retrieval approach in autonomous driving. 
At present, we primarily focus on the image representation of scenes. However, our framework can be expanded to incorporate other sensors, such as LiDAR or a fusion of LiDAR and cameras, which can be achieved by merely substituting the corresponding BEV encoder\cite{bai2022transfusion, liang2022bevfusion}.
% In our current implementation, 
Currently, we employ pre-trained BEV and fine-tuned language encoders to effectively perform retrieval tasks. It would be valuable to explore the potential of leveraging our framework for pre-training tasks, similar to the pioneering CLIP~\cite{radford2021learning}.
Additionally, in this study, we conducted experiments with various efficient learning methods for KGE. In the future, more sophisticated networks, e.g., Graph Neural Networks (GNN)~\cite{wu2020comprehensive}, could be introduced to enhance the learning of embeddings.
We hope that our work can inspire future research in this field.

\section{Additional Details}

\mypara{Knowledge Graph in autonomous driving.}
Knowledge graphs (KGs) are known for exploiting structured, dynamic, and relational data~\cite{cao2024knowledge, wang2017knowledge}. 
The autonomous driving industry is also exploring the use of knowledge graphs to manage the vast amount of heterogeneous data, and researchers are actively investigating the potential benefit of KGs applied to autonomous driving tasks, e.g., perception, scene understanding, and motion planning~\cite{wormann2022knowledge, luettin2022survey}.
ADKG~\cite{nag2021towards} studied the role of commonsense knowledge in autonomous driving applications and constructed a large-scale autonomous driving knowledge graph. 
Wickramarachchi et al.~\cite{wickramarachchi2020evaluation} focused on embedding AD data and investigated the quality of the trained embeddings given various degrees of AD scene details within the KG.
In this work, we first explore the utilization of autonomous driving knowledge graphs for text-scene retrieval. As traffic scenes involve complex relationships that are not easily expressed through textual input alone, we leverage the autonomous driving knowledge graph to enhance the semantic representation of the encoded text.

\mypara{Datasets.}
For the retrieval dataset, we adopt the \nusr{} in \cref{sec:nusr} based on the nuScenes dataset~\cite{caesar2020nuscenes}, which is a large-scale public dataset for autonomous driving. It contains a total of 1,000 driving scenarios collected in Boston and Singapore, with each scenario lasting about 20 seconds. Each sample is captured by RGB cameras distributed in six surrounding viewpoints of the car and a LiDAR placed on the roof. The training and validation sets contain a total of 34,149 labeled keyframes.
For the knowledge graph data, we utilize the autonomous driving knowledge graph (ADKG)~\cite{nag2021towards} to train knowledge graph embedding for keywords in autonomous driving. ADKG is generated based on perceptual data from Hesai's PandaSet~\cite{xiao2021pandaset}. It contains more than 57,000 instances and more than 330,000 triplets. It also contains 7 action labels and more than 40 object labels. 
Based on these labels, we manually perform synonym mapping so that the caption mentioned above can trigger the knowledge graph retrieval more frequently and accurately.

\mypara{Metrics.}
We use (R@K, K=1,5,10) as our evaluation metrics of recall accuracy, which is the most commonly used evaluation metric in the retrieval tasks and is the abbreviation for recall at $k$-th in the ranking list, defined as the proportion of correct matchings in top-$k$ retrieved results. 
% In the following experimental results, B2T and T2B refer to BEV-to-text retrieval and text-to-BEV retrieval, respectively.

\mypara{Implementation details.}
In our experiments, the default size of the BEV feature is (2500,1,256), and the hidden size of the language embedding is set to 4096. 
The dimensions of the knowledge graph embeddings and shared cross-modal embeddings are both 1024, respectively. 
All experiments are conducted on 8 NVIDIA A100-80GB GPUs.
We optimize the knowledge graph embeddings using SGD optimizer with initial learning rate of 0.25 with 16k iterations until convergence, and train our \name{} for 80 epochs with a batch size of 128 by AdamW optimizer~\cite{loshchilov2017decoupled} with initial learning of 1e-4 adjusted by cosine annealing strategy~\cite{loshchilov2016sgdr}.
The weight balance coefficient $\lambda$ is set to 0.15 by default.
Without any specifications, the default BEV encoder of \name{} is the BEVFormer-base~\cite{li2022bevformer} encoder, and the default text encoder is Llama2~\cite{touvron2023llama} with LoRA~\cite{hu2021lora} fine-tuning strategy.
% consisting of a ResNet101-DCN [11, 19] backbone with an FPN neck [51] and additional 6 encoder layers to extract BEV features from multi-view image sequences.
% The baseline text encoder we use is based on the BERT model, which is natively used in CLIP and uses an MLP layer for feature mapping. 
% For the baseline of the experiments, we employ the CLIP native text branch as our adaptive method and replace the original visual branch with the BEVFormer encoder. 
% Additionally, we insert a layer of Multilayer Perceptron (MLP) between the BEV branch and contrastive loss to align the feature size.

\section{Additional Results}
% \subsection{Ablation studies on \nusqar{}}
% \input{latex/tab/ablation_nusqa}
% \input{latex/tab/ablation_kg_2}
% \input{latex/tab/ablation_caption_2}
% Here we provide additional ablation experiment results for our augmented NuScenes-QA data. Table \ref{table:table 10} refers results when using different knowledge graph. Table \ref{table:table 12} refers results of adding caption generation as an auxiliary task. These results all agree with the effective improvement of applying components that we proposed in this paper, and verified the abilities of zero-shot corner case retrieval abilities for our method.
\begin{table*}[h]
  \centering 
   \addtolength{\tabcolsep}{1.8pt}
    \begin{tabularx}{0.8\linewidth}{c|ccc|ccc}
\toprule 

   % Module                     & B2T\_R1        & B2T\_R5        & B2T\_R10       & T2B\_R1         & T2B\_R5         & T2B\_R10        \\
   \multirow{2}{*}{Dimension}  & \multicolumn{3}{c|}{Text Retrieval} & \multicolumn{3}{c}{Scene Retireval}  \\
      \cmidrule(r){2-7}   & R@1 & R@5 & R@10  & R@1 & R@5 & R@10 \\

%               \multicolumn{3}{l}{\textit{\textbf{
% Shared Cross-modal Embedding}  }}  \\
    \midrule
   $d$ =    512   & 0.8386 & 0.9885 & 0.9986          & 0.8753 & 0.9914 & 0.9986          \\
    \rowcolor{mygray} $d$ =  1024  & 0.8578 & 0.9954 & 0.9994          & 0.8766 & 0.9971 & 0.9997          \\
      $d$ = 2048 & 0.8530 & 0.9942 & 0.9986          & 0.8862 & 0.9975 & 0.9997          \\

\bottomrule 
\end{tabularx}
\caption{\textbf{Ablations on the dimension of Shared Cross-modal Embedding}.}
\label{tab: ablation_supp}
\end{table*}
% baseline + KG
\subsection{Ablations on the dimension of Shared Cross-modal Embedding}
In \cref{tab: ablation_supp}, we investigate the impact of the feature dimension of the shared cross-modal embeddings. The results indicate that a larger dimension results in improved retrieval performance as it enables the effective alignment of features from both modalities. Moreover, a dimension of $d = 1024$ is sufficient to generate favorable retrieval results.

\section{Additional Visualizations}
\begin{figure}[ht]
        \centering
        \includegraphics[width=1\linewidth]{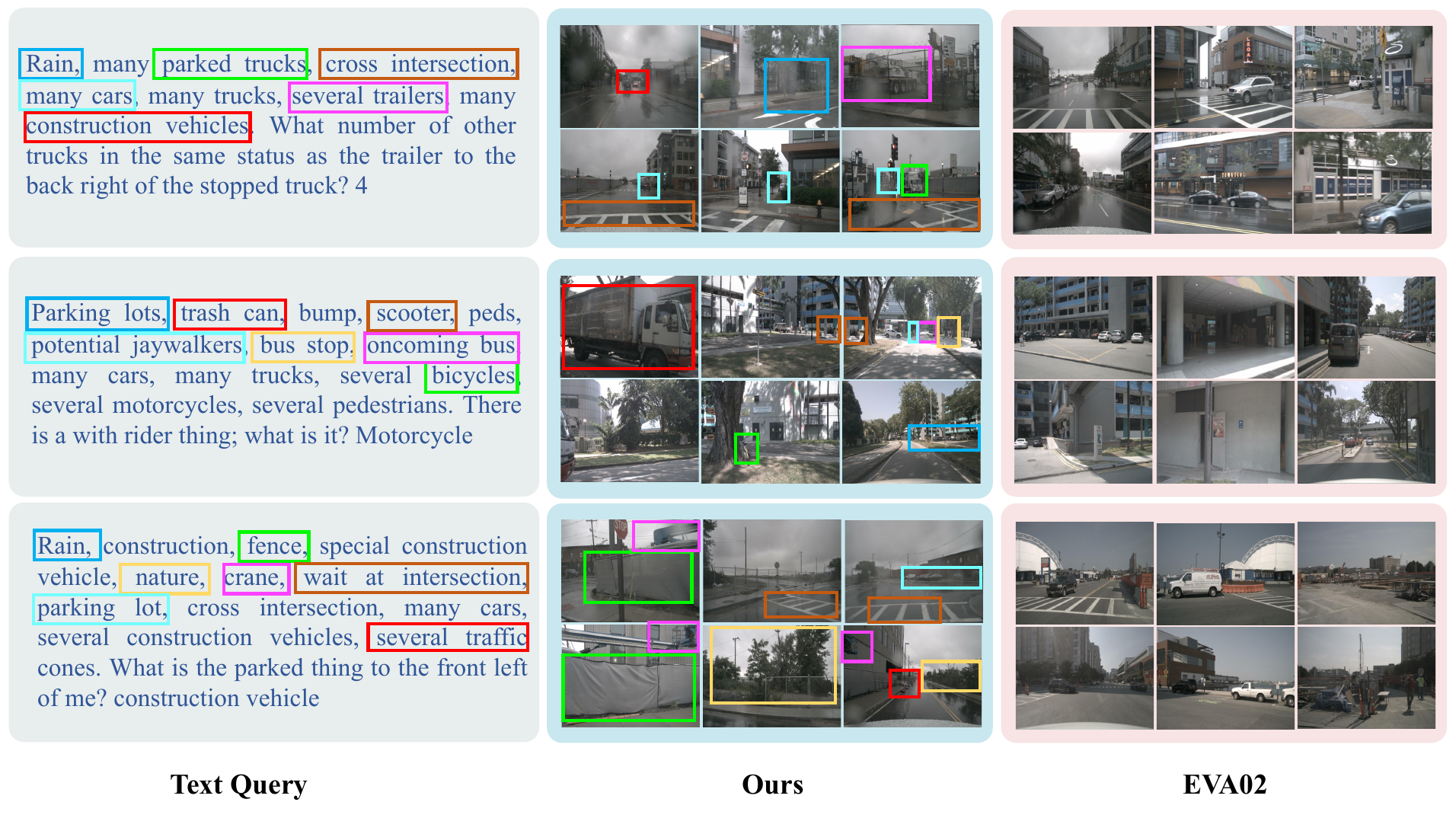}
        \caption{\textbf{R@1 scene retrieval results.} Our \name{} achieves more precise retrieval results.}
        \label{fig:vis_supp}
\end{figure}
In \cref{fig:vis_supp}, we present additional qualitative results obtained from the \nusr{} dataset. We compare our proposed method, \name{}, with the previous state-of-the-art retrieval method EVA02~\cite{fang2023eva}.
EVA02, being an image space retrieval model, struggles to find scenes that correspond well to the textual queries. In contrast, our \name{} achieves more precise retrieval results. It demonstrates a significant capability to understand the global context. Furthermore, even under challenging weather conditions, such as a rainy day, our \name{} achieves satisfactory retrieval results.

\end{document}